\documentclass[sigconf]{acmart}
\usepackage{setspace}
\usepackage{multirow}

\AtBeginDocument{%
  \providecommand\BibTeX{{%
    \normalfont B\kern-0.5em{\scshape i\kern-0.25em b}\kern-0.8em\TeX}}}

\copyrightyear{2022}
\acmYear{2022}
\setcopyright{acmcopyright}\acmConference[WWW '22]{Proceedings of the ACM Web Conference 2022}{April 25--29, 2022}{Virtual Event, Lyon, France}
\acmBooktitle{Proceedings of the ACM Web Conference 2022 (WWW '22), April 25--29, 2022, Virtual Event, Lyon, France}
\acmPrice{15.00}
\acmDOI{10.1145/3485447.3511935}
\acmISBN{978-1-4503-9096-5/22/04}

\newcommand{\TaxoExpan}{\mbox{\sf TaxoExpan }}
\newcommand{\GenTaxo}{\mbox{\sf GenTaxo }}
\newcommand{\TaxoEnrich}{\mbox{\sf TaxoEnrich }}

\newcommand{\TaxoEnrichS}{\mbox{\sf TaxoEnrich-S }}
\newcommand{\TMN}{\mbox{\sf TMN }}

\newcommand{\mquote}[1]{{``\emph{#1}''}}

\begin{document}

\title{TaxoEnrich: Self-Supervised Taxonomy Completion via Structure-Semantic Representations}

\author{Minhao Jiang$^{1}$, Xiangchen Song$^{2}$, Jieyu Zhang$^{3}$, Jiawei Han$^{1}$}
\affiliation{%
$^{1}$Department of Computer Science, University of Illinois Urbana-Champaign, IL, USA\\
$^{2}$School of Computer Science, Carnegie Mellon University, Pittsburgh, PA, USA\\
$^{3}$The Paul G. Allen School of Computer Science \& Engineering, University of Washington, WA, USA
\country{}
}
\email{ {minhaoj2, hanj}@illinois.edu, xiangchs@cs.cmu.edu, jieyuz2@cs.washington.edu}
\renewcommand{\shortauthors}{Jiang et al.}


\begin{abstract}
Taxonomies are fundamental to many real-world applications in various domains, serving as structural representations of knowledge. To deal with the increasing volume of new concepts needed to be organized as taxonomies, researchers turn to automatically completion of an existing taxonomy with new concepts. 
In this paper, we propose {\sf TaxoEnrich}, a new taxonomy completion framework, which effectively leverages both semantic features and structural information in the existing taxonomy and offers a better representation of candidate position to boost the performance of taxonomy completion.
Specifically, \TaxoEnrich consists of four components:
(1) taxonomy-contextualized embedding which incorporates both semantic meanings of concept and taxonomic relations based on powerful pretrained language models;
(2) a taxonomy-aware sequential encoder which learns candidate position representations by encoding the structural information of taxonomy;
(3) a query-aware sibling encoder which adaptively aggregates candidate siblings to augment candidate position representations based on their importance to the query-position matching;
(4) a query-position matching model which extends existing work with our new candidate position representations.
 Extensive experiments on four large real-world datasets from different domains show that \TaxoEnrich achieves the best performance among all evaluation metrics and outperforms previous state-of-the-art methods by a large margin.
\end{abstract}

\ccsdesc[500]{Computing methodologies~Information extraction}

\keywords{Taxonomy Completion, Self-supervised Learning, Knowledge Representation}

\maketitle
\begin{spacing}{0.97}
\begin{figure}[b]
\centering
\includegraphics[width=0.5\textwidth]{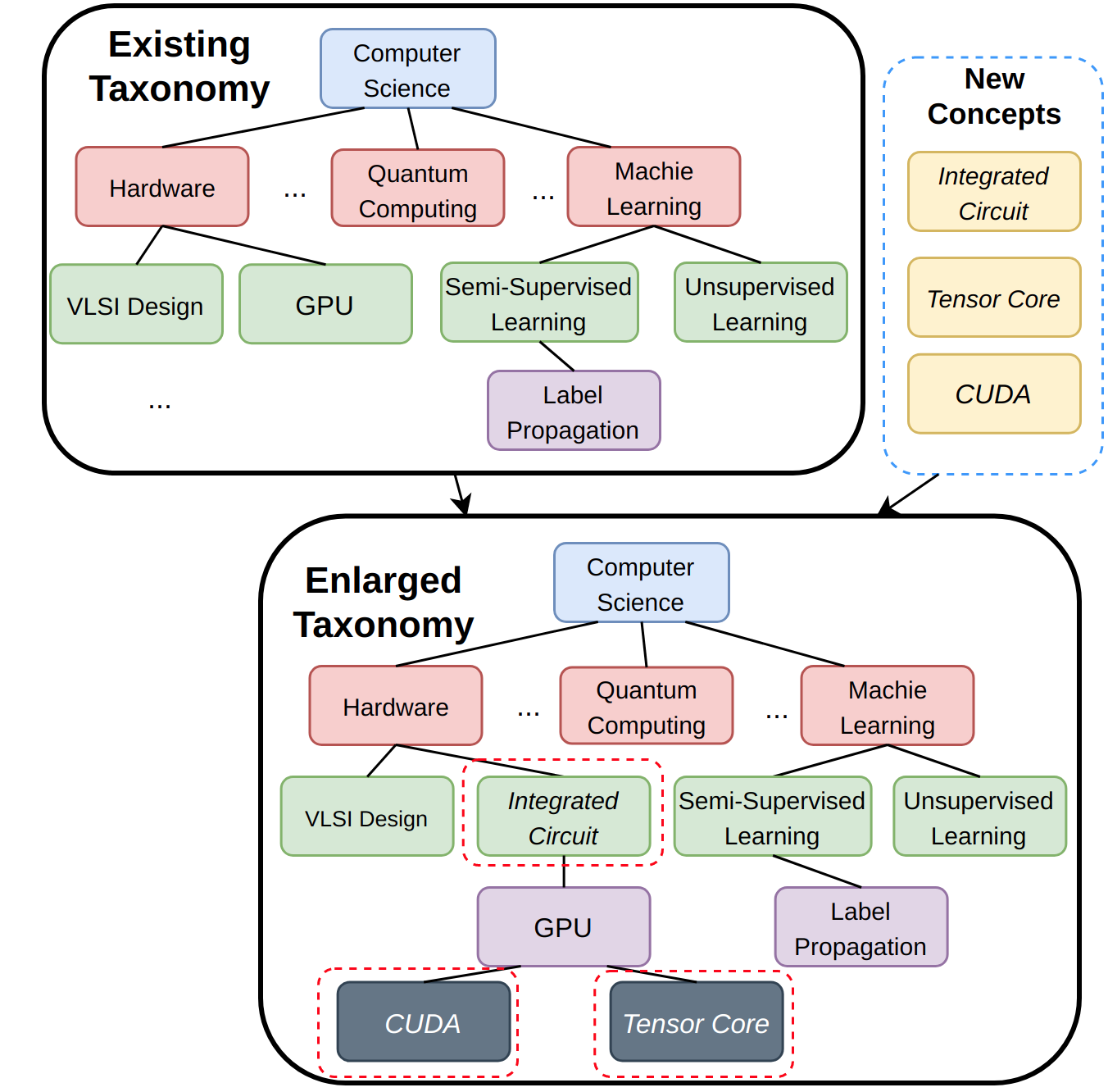}
\caption{An example of inserting new concepts into an existing taxonomy of computer science terms. For each new concept, we aim to find the relatedness between the concept and each candidate position.}
\label{fig:intro}
\end{figure}

\section{Introduction}
Taxonomies have been widely used for organizing concepts structurally \cite{vrande2012, shen2018entity, sinha2015overview, mao2020, yang2020co}. Specifically, to capture the ``is-a'' relationship between concept pairs, people often formulate the taxonomy into tree or directed acyclic graph (DAG) structure.
Example applications could be found in e-commerce, where Amazon leverages product taxonomies for personalized recommendations and product navigation, and fine-grained named entity recognition where people rely on concept taxonomies (e.g., MeSH) \cite{lipscomb2000medical} to extract and label useful information from massive corpus.

However, the construction of taxonomies usually requires a substantial amount of human curation. Such process is time-consuming and labor-intensive.
Thus, it is extremely hard to handle the large number of emerging new concepts in downstream tasks, which is fairly common nowadays with the rising tide of big data. To tackle this issue, recent work \cite{shen2020taxoexpan, manzoor2020expanding,yu2020steam, zhang2021taxonomy, zeng2021enhancing, song2021should} turns to the tasks of the automatic expansion and completion of the existing taxonomy.

Previous taxonomy construction methods \cite{gupta2017taxonomy, mao2018end, zhang2018taxogen, liu2012automatic, wang2013phrase, hearst1992automatic} construct taxonomies from scratch, and highly rely on annotated hypernym pairs, which are expensive and sometimes inaccessible in practice. Therefore, automatic taxonomy expansion based on existing taxonomies is in great need and has gained increasing attention. 

The recent studies on taxonomy expansion and completion achieved noticeable progress, which mainly contribute from two directions. (1) extract hierarchical information from the existing taxonomy, and utilize different ways to model the structural information in the existing taxonomy, such as local egonet \cite{shen2020taxoexpan}, parent-query-child triplet \cite{zhang2021taxonomy}, and mini-paths \cite{yu2020steam}. (2) leverage the supporting corpus to generate the embeddings of concepts directly.  They either only used implicit relational semantics \cite{manzoor2020expanding}, or only relied on corpus to construct limited seed-guided taxonomy \cite{huang2020corel}. Very recently, \cite{zeng2021enhancing} combines the representations from semantic sentences and local-subgraph encoding as the features of concepts. However, they only utilized light-weight multi-layer perceptron (MLP) for matching, which suffers from the limited representation power. In this paper, we follow \cite{zhang2021taxonomy} to focus on taxonomy completion, which aims to predict the most likely $\langle$query, hypernym, hyponym$\rangle$ triplet for a given query concept. For example, in Figure \ref{fig:intro}, when considering the query \mquote{Integrated Circuit}, we aim to find its true parent \mquote{Hardware} and child  \mquote{GPU}.

To effectively leverage both semantic and structural information for better taxonomy completion performance, in this work, we propose {\sf TaxoEnrich}, which aims to learn better representations for each candidate position and render new state-of-the-art taxonomy completion performance.
Specifically, \TaxoEnrich consists of four carefully-designed components.
First, we propose a taxonomy-contextualized embedding generation process based on pseudo sentences extracted from existing taxonomy. 
The two types of pseudo sentences, i.e., ancestral and descendant pseudo sentences, capture taxonomic relations from two directions respectively.
Then, the powerful pretrained language models are utilized to produce the taxonomy-contextualized embedding based on the extracted sentences. 
Secondly, to encode the structural information of the existing taxonomy in both vertical and horizontal views, we develop two novel encoders: a sequential feature encoder based on the pseudo sentences and a query-aware sibling encoder base on the importance of candidate siblings to the matching task.
The former aims to learn a taxonomy-aware candidate position representations, while the latter further augments the position representations with adaptively aggregated candidate siblings information.
Finally, we develop an effective query-position matching model by extending previous work~\cite{zhang2021taxonomy} to incorporate our novel candidate position representations.
Specifically, it takes into consideration both fine-grained (query to candidate parent) and coarse-grained (query to candidate position) relatedness for better taxonomy completion performance.

We conducted extensive experiments on four real-world taxonomies from different domains to test the performance of {\sf TaxoEnrich} framework. Further more, we designed two variations of the framework, \TaxoEnrich and {\sf TaxoEnrich-S} to conduct ablation experiments to explore the utilization of different information under different datasets, along with studies to examine the effectiveness of each sub-module of the framework. Our results show that \TaxoEnrich can more accurately capture the correct positions of query nodes than previous methods and achieve state-of-the-art performance on both taxonomy completion and expansion tasks.

To summarize, our major contributions include:
\begin{itemize}
    \item We propose an effective embedding generation approach which can be applied generally for learning contextualized embedding for each concept node in a given taxonomy.
    \item We introduce the sequential feature encoders to capture vertical structural information of candidate positions in the taxonomy.
    \item We design an effective query-aware sibling encoder to incorporate horizontal structural information in the taxonomy.
    \item Extensive experiments demonstrate that our developed framework enhances the performance on both taxonomy completion and expansion task by a large margin over the previous works. 
\end{itemize}

\begin{figure*}
    \centering
    \includegraphics[width=\textwidth]{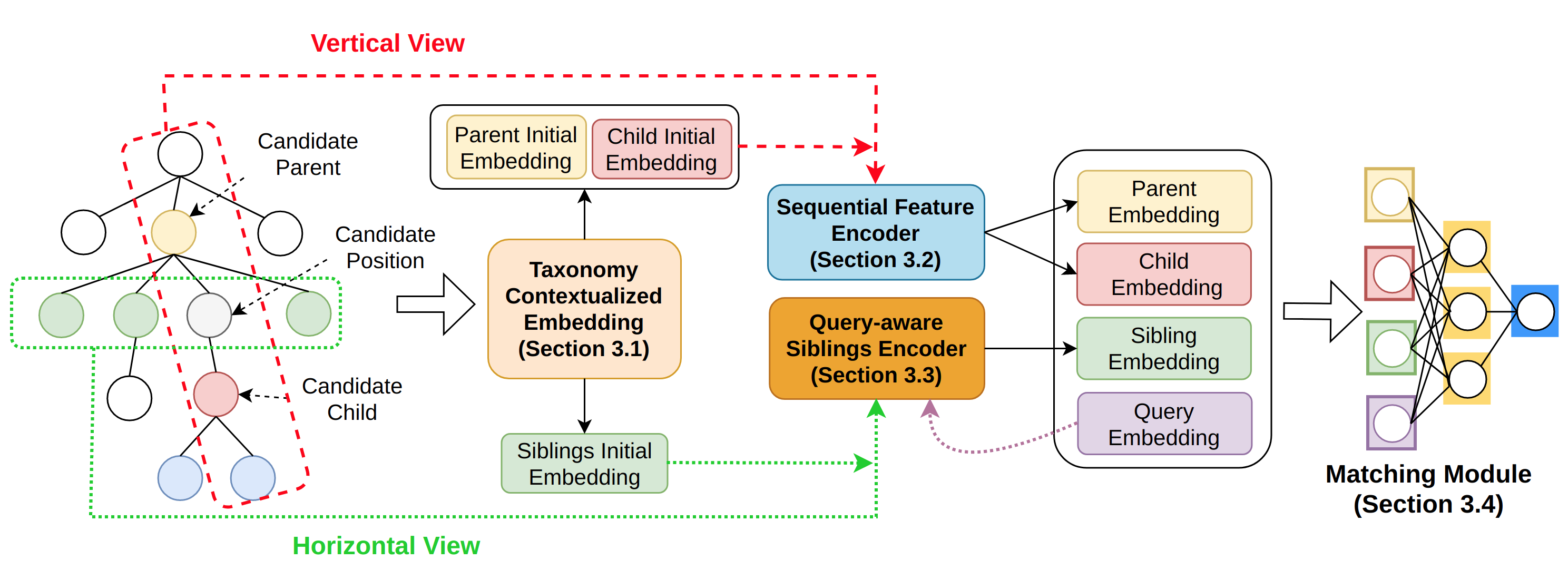}
    \caption{The complete architecture of \TaxoEnrich. The figure describes the workflow of {\sf TaxoEnrich}, and the details are discussed in the corresponding section.}
    \label{fig:framework}
\end{figure*}

\section{Problem Definition}
In this section, we formally define the taxonomy completion task studied in the paper. 

\noindent \textbf{Taxonomy.} Follow \cite{shen2020taxoexpan}, we formulate a taxonomy $\mathcal{T}^0 = \{\mathcal{N}^0, \mathcal{E}^0\}$ as a directed acyclic graph where each node $n \in \mathcal{N}^0$ represents a concept and each directed edge $\langle n_p, n_c \rangle \in \mathcal{E}^0$ represents a taxonomic relationship between two concepts. 

\noindent \textbf{Taxonomy Completion.} The taxonomy completion task~\cite{zhang2021taxonomy} is defined as following: given an existing taxonomy $\mathcal{T}^0$ and a set of new concepts $\mathcal{C}$, assuming that each concept in $\mathcal{C}$ is in the same semantic domain as $\mathcal{T}^0$, we aim to automatically find the most possible $\langle$ hypernym, hyponym$\rangle$ pairs for each new concept to complete the taxonomy. The output is $\mathcal{T} = \{\mathcal{N}, \mathcal{E^0}\}$ where $\mathcal{N} = \mathcal{N}^0 \cup \mathcal{C}$ and $\mathcal{E^0}$ is the updated edges set after inserting new concepts.

\noindent \textbf{Candidate Positions. } In the taxonomy completion task, we define a valid candidate position as a pair of concept nodes in the existing taxonomy $\langle n_p, n_c \rangle$ where $n_p$ is a parent of $n_c$. 
Note that one of $n_p$ and $n_c$ could be a pseudo placeholder node in case that the concepts needed to be inserted as root or leaf nodes. 

The goal of taxonomy completion is to enrich the existing taxonomy by inserting new concepts. These new concepts are generally extracted from text corpus using entity extraction tools. Since this process is not the focus of the paper, we assume that the set of new concepts $\mathcal{C}$ is given, as well as their embedding, which is denoted by $e_q$ for new concept $n_q$.

\section{The TaxoEnrich Framework}
In this section, we introduce the \TaxoEnrich framework in details.
We first introduce the taxonomy-contextualized embedding generation for each concept node in the existing taxonomy. 
Then, given the extracted taxonomy-contextualized embedding, we develop two encoders to learn the representation of candidate positions from vertical and horizontal views of the taxonomy respectively. 
Finally, we propose a query-to-position matching model which leverages various structural information and takes into consideration both fine- and coarse-grained relatedness to boost the matching performance. The overall framework of \TaxoEnrich is in Figure~\ref{fig:framework}.

\subsection{Taxonomy-Contextualized Embedding}
\label{31}
\begin{figure}[b]
    \centering
    \includegraphics[width=0.5\textwidth]{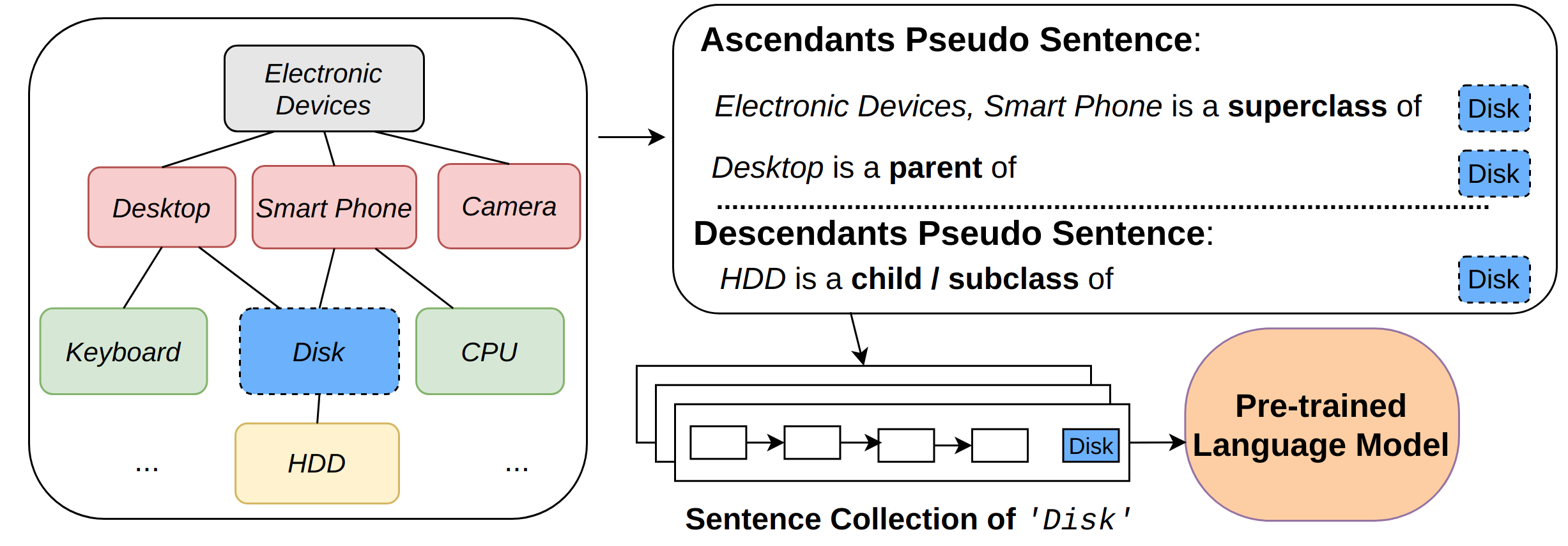}
    \caption{The visualization of taxonomy-contextualized generation process. In this example, we aim to extract the embedding of the concept node \mquote{Disk}.}
    \label{fig3}
\end{figure}

Here, we describe the generation process of taxonomy-contextualized embedding for each node in the taxonomy.
Different from prior work which leverages static word embedding, such as Word2Vec and FastText \cite{shen2020taxoexpan, zhang2021taxonomy}, or contextualized embedding solely based on an additional text corpus \cite{yu2020steam}, we generate taxonomy-contextualized embedding based on taxonomy structure and concept surface name.
The reason is that neighboring concepts in taxonomy are likely to share similar semantic meaning and it is hard to distinguish them based on predefined general-purpose embedding.
With similar spirit, \cite{zeng2021enhancing} also leverage pretrained language models to produce contextualized embedding based on limited number of taxonomy neighbor and the surface names is implicitly utilized in fine-tuning the pretrained language models.
In contrast, we aim to fuse the information of all the descendant/ancestral concepts of the given concept, without fine-tuning a huge pretrained language model.
Specifically, given a concept node, we build pseudo sentences based on Hearst patterns \cite{roller-etal-2018-hearst} to represent both the positional and semantic information.
We separately consider the descendant/ancestral information by constructing descendant/ancestral pseudo sentences respectively  as shown in Figure \ref{fig3}. 

Formally, given a taxonomy $\mathcal{T}_0 = (\mathcal{V}_0, \mathcal{E}_0)$ and the candidate concept node $v$, we extract the following two types of pseudo sentences that represent taxonomic relationships:
\begin{enumerate}
    \item Ancestral Pseudo Sentences: We first extract paths connecting root and the candidate node $v$ without duplicate nodes. In the extracted path $(p_1, p_2, \ldots, p_l)$, $p_1$ is the root node and $p_l = v$. We denote the $i$-th extracted path for node $v$ as $P_{a}(v)_i = (p_1, p_2, \ldots, p_{l-1})$. Along each path, we generate the ancestral pseudo sentence  as below:
    \begin{center}
        "$p_1, p_2, \ldots, p_{l-1}$ is a superclass of $v$"
        \\
        or "$p_1, p_2, \ldots, p_{l-1}$ is an ascendant of $v$"
    \end{center}
    
    \par All such words like "superclass" or "ascendant" that can represent the hierarchical relationship between the path and candidate nodes can be used for sentence generation. We denote the collection of such ancestral paths as $P_a(v)$ and the generated sentences as $S_a(v)$
    \item Descendant Pseudo Sentences: Similarly, the paths without duplicated nodes starting from the candidate node $v$ to leaf nodes are extracted, denoted as $(p_1, p_2, \ldots, p_l)$ where $p_1 = v$ and $p_l$ is a leaf node. In this case, along each path, we generate the sentence as below
    \begin{center}
        "$p_2, p_3, \ldots, p_{l}$ is a subclass of $v$"
        \\
        or "$p_2, p_3, \ldots, p_{l}$ is a descendant of $v$"
    \end{center}
    We denote the collection of such descendants paths as $P_d(v)$ and the generated sentences as $S_d(v)$
\end{enumerate}
Then, the set of all the generated pseudo sentences is $S(v) = S_d(v) \cup S_a(v)$. As visualized in Figure \ref{fig3}, aiming to generate embeddings of the concept node \mquote{Disk}, we only consider the ancestral and descendant paths, such as \mquote{Electronic Devices, Smart Phone is a superclass of Disk}. Note that if the candidate node is leaf node or root, we will only consider one side pseudo sentences. Given the generated pseudo sentences, we apply a pretrained language models to generate taxonomy-contextualized embedding for each node. Specifically, we feed the pseudo sentences to the pretrained language model and collect the last hidden state representations of the concept node $v$, which is averaged as the final taxonomy-contextualized embedding $\mathbf{x}_v \in \mathbb{R}^d$.
In our preliminary experiments, we found that SciBERT is better than other models in representing concept representations. Hence, in this paper, we choose SciBERT \cite{beltagy2019scibert} following~\cite{zeng2021enhancing}. And the comparison between different pre-trained language models in terms of performance are discussed in \ref{appendix}. 
Note that the taxonomy-contextualized embeddings are pre-computed and fixed for the following modules, which means we do not fine-tune the large pretrained language model.

\subsection{Sequential Feature Encoder}

Given the taxonomy-contextualized embedding $\mathbf{x}_v$ for existing node $v$, we develop a learnable sequential feature encoder $g(v)$ to encode the structural information of candidate positions in a vertical view of taxonomy.
For a candidate position $\langle p, c \rangle$ consisting of candidate parent $p$ and child $c$, we produce parent embedding $g(p)$ and child embedding $g(c)$ respectively.
Specifically, for candidate parent $p$ and its corresponding ancestral paths $P_{a}(p)$,
we randomly sample a path $p_p$ from $P_{a}(p)$ and apply a LSTM sequential encoder which inputs the sampled pseudo sentence and the taxonomy-contextualized embedding.
Then, we concatenate the final hidden state of the LSTM encoder and the taxonomy-contextualized embedding $\mathbf{x}_p$ as parent embedding.
Formally,
\begin{equation}
    g(p) = \mathbf{x}_p \oplus \text{LSTM}(p_p; \Theta_1)
\end{equation}
where $\Theta_1$ is the learnable parameters of the LSTM encoder and $\oplus$ represents the concatenation operation.
Similarly, we generate the child embedding $g(c)$ based on taxonomy-contextualized embedding $\mathbf{x}_c$ and descendant paths $p_c$ from $P_{d}(c)$:
\begin{equation}
    g(c) = \mathbf{x}_c \oplus \text{LSTM}(p_c; \Theta_2)
\end{equation}
where, $\Theta_2$ represents the learnable parameters of the LSTM encoder. The output $g(u), g(v) \in \mathbb{R}^h$ will be used as the embedding for candidate position nodes.

Through this sequential feature encoder, we are able to fuse the structural information of candidate position in a vertical view.
This allows the candidate position representations to be aware of the “depth" information of the candidate position, i.e., whether the candidate position is in the top-level of taxonomy close to the root or in the bottom-level close to leave.

\subsection{Query-Aware Siblings Encoder}

The aforementioned sequential feature encoder incorporates the taxonomy structural information in a vertical view, however, it is of great importance to also encode the horizontal local information of the candidate position.
Thus, we develop another encoder to incorporate the structural information in a horizontal view.
Specifically, in addition to the candidate parent and child, we consider the candidate siblings, i.e., the children of candidate parent, of the query node.

However, incorporating candidate siblings is challenging than candidate parent and child.
The reasons are twofold. First, compared to the candidate parent and child which compose the candidate position, candidate siblings could introduce noisy information and thus lead to sub-optimal results. For example, for the top-level of taxonomy, the candidate siblings could have quite diverse semantic meanings, which hinder good matching between candidate position and query node.
Secondly, since some candidate parent could have substantial amount of children (candidate siblings), it is infeasible to incorporate all the candidate siblings without strategic selection.

To tackle these issues, we develop a query-aware siblings encoder, which adaptively selects part of the candidate siblings.
Specifically, we measure the relatedness of a given query embedding $e_q$ and each candidate sibling condition on the representation of candidate parent-child pair.
Such relatedness is in turn used to aggregate the sibling information into a single siblings embedding.
Mathematically, given candidate position $\langle n_p, n_c \rangle$ with corresponding embedding $g(p), g(c)$ and the set of candidate siblings $C(n_p)=\{s_1, s_2, \ldots, s_t\}$, we use a learnable bilinear matrix $\mathbf{W}_{\text{Sib}} \in \mathbb{R}^{d \times (2h+d)} $ to calculate the relatedness of query and candidate sibling $s_i$ as
\begin{equation}
    \phi_{s_i} = e_q^T\mathbf{W}_{\text{Sib}} \begin{bmatrix} g(p), g(c), \mathbf{x}_{s_i} \end{bmatrix}
\end{equation}

Then the relatedness score $\phi_{s_i}$ is normalized over the set of candidate siblings by a softmax function:
\begin{equation}
    \alpha_{s_i} = \sigma_{\text{softmax}}(\phi_{s_i}) = \frac{\text{exp}(\phi_{s_i})}{ \sum_{s_j \in C(n_p)} \text{exp}(\phi_{s_j})}
\end{equation}

The normalized score $\alpha_{s_i}$ captures the importance of candidate sibling $s_i$ for the specific query-position matching. 
In other words, it highlights the siblings relevant to the query condition on the candidate position while lessen the effect of irrelevant siblings.
Finally, the sibling embeddings are aggregate based on the normalized score as 
\begin{equation}
    a(p) = \sum_{s_i \in C(n_p)} \alpha_{s_i} \mathbf{x}_{s_i}
\end{equation}

where $a(p) \in \mathbb{R}^d$. During experiments, we found that such a query-aware siblings encoder renders good performance when only a subset of siblings are considered, which alleviates the heavy burden of aggregate over the potentially large amount of candidate siblings.

\subsection{Query-Position Matching Model}

Finally, given the representation of candidate parent $g(p)$, child $g(c)$ and siblings $a(p)$ as well as the given query embedding $e_q$, we are ready to present our final matching module, which outputs the matching score of query and candidate position for taxonomy completion task.
In particular, we seek to learn a matching model $s$ that outputs the desired relatedness score: 
\begin{equation}
    s(n_q, \langle n_p, n_c \rangle) = f(e_q, g(p), g(c), a(p))
\end{equation}
where $f$ is a parametrized scoring function. 

The previous study \cite{zhang2021taxonomy} showed that the simple matching model that learns one-to-one relatedness between the query node and the position pair ignores fine-grained relatedness between query and position component, i.e., the relatedness between $\langle n_q, n_p \rangle$ and $\langle n_q, n_c \rangle$. Therefore, inspired by \cite{zhang2021taxonomy}, we propose a new matching model which incorporates the additional siblings  embedding and learn more precise matching based on both fine-grained (query to candidate parent/child/siblings) and coarse-grained relatedness (query to position).

To learn both the fine-grained and coarse-grained relatedness between the query node and the candidate positions, we construct multiple auxiliary scorers that separately focus on the relationship between the query node and the candidate parent, the candidate child, the candidate siblings and the candidate position, respectively.
We adopt the Neural Tensor Network (NTN) \cite{socher2013reasoning} as the base models. Given vectors $u\in\mathbb{R}^{d_u}, v\in\mathbb{R}^{d_v}$, an NTN can be defined as 
\begin{equation}
    \text{NTN}(u, v) = \mathbf{w}^{T} \sigma_{\text{tanh}}(h(u, v))
\end{equation}
\begin{equation}
    h(u, v) = u^T \mathbf{W}^{[1:k]} v + \mathbf{V} \begin{bmatrix} u \\ v \end{bmatrix} + \mathbf{b}
\end{equation}
where $\sigma_{\text{tanh}}$ is a tanh function and $\mathbf{w} \in \mathbb{R}^{k}$, $\mathbf{W}^{[1:k]} \in \mathbb{R}^{d_{u} \times d_{v} \times k}$, $\mathbf{V} \in \mathbb{R}^{k \times (d_{u}+d_{v})}$ and $\mathbf{b} \in \mathbb{R}^k$ are learnable parameters. Note that $k$ is a hyperparameter in NTN.

Then our multiple scorer can be defined as 
\begin{equation}
    S_1(n_q, n_p) = \mathbf{w}_1^{T} \sigma_{\text{tanh}}(h_1(e_q, {g}(p)))
\end{equation}
\begin{equation}
    S_2(n_q, n_c) = \mathbf{w}_2^{T} \sigma_{\text{tanh}}(h_2(e_q, {g}(c)))
\end{equation}
\begin{equation}
    S_3(n_q, C(n_p)) = \mathbf{w}_3^{T} \sigma_{\text{tanh}}(h_3(e_q, {a}(p)))
\end{equation}
\begin{equation}
    S_4(n_q,  \langle n_p, n_c \rangle) = \mathbf{w}_4^{T} \sigma_{\text{tanh}}(h_4(e_q,  \begin{bmatrix} {g}(p), {g}(c), {a}(p) \end{bmatrix}))
\end{equation}
We omit the learnable parameters inside each $h$ for notation convenience.
In this formulation, $S_1, S_2, S_3$ aim to learn the fine-grained relatedness for $\langle n_q, n_p\rangle, \langle n_q, n_c\rangle, \langle n_q, C(n_p)\rangle$ separately by predicting whether the $n_p$, $n_c$, and $C(n_p)$ is the reasonable parent, child, and siblings, respectively. 
Differently, $S_4$ is designed for coarse-grained relatedness between the query node and the candidate position. 
Eventually we construct a primal scorer which incorporates the all the auxiliary scorers.
\begin{equation}
    S_p(n_q,  \langle n_p, n_c \rangle) = \mathbf{u}_p^T \sigma_{\text{tanh}}(\begin{bmatrix} h_1, h_2, h_3, h_4 \end{bmatrix}))
\end{equation}
We omit the input of each function $h$ for simplicity.
In this case, even though $S_p$ and $S_4$ share the same supervision signal, the concatenation of internal representations of other auxiliary scorers in $S_p$ will allow it to capture accurate matching information based on $S_1, S_2, S_3$ when $S_4$ cannot learn correct coarse-grained relatedness.

\subsubsection{Learning Objectives}
For each auxiliary scorers, since the model is trained for binary classification task to calculate the relatedness between the query node and the target objective, we adopt the binary cross-entropy loss. Thus, the learning objective for each scorer can be formulated as
\begin{equation}
    \mathcal{L}_k = - \frac{1}{|\mathbb{D}|} \sum_{(\mathbf{X}_i, y_i) \in \mathbb{D}} y_i \cdot \log (S_i(\mathbf{X}_i)) + (1 - y_i) \cdot \log (1 - S_i(\mathbf{X}_i))
\end{equation}
where $\mathbb{D}$ is the dataset formulated by the self-supervised generation following similar methods proposed in \cite{shen2020taxoexpan, zhang2021taxonomy}, and $(\mathbf{X}_i, y_i)$ is the generated data pair in the dataset, and $k \in \{p, 1, 2, 3, 4\}$ represents each scorer. In this case, the final learning objective $\mathcal{L}(\Theta)$ that focuses on the primal task will naturally be defined as 
\begin{equation}
    \mathcal{L}(\Theta) = \mathcal{L}_p + \lambda_1 \mathcal{L}_1 + \lambda_2 \mathcal{L}_2 + \lambda_3 \mathcal{L}_3 + \lambda_4 \mathcal{L}_4
\end{equation}

\begin{table}[t]
    \caption{Dataset Statistics. $|\mathcal{N}|$ represents the number of nodes in the taxonomy and $|\mathcal{E}|$ represents the number of edges in the taxonomy. $|\mathcal{D}|$ indicates the taxonomy depth. \# of Sentences denotes the number of pseudo sentences generated by the embedding generation module in each taxonomy.}
    \begin{tabular}{c c c c c}
    \hline
    \textbf{Dataset} & $|\mathcal{N}|$ & $|\mathcal{E}|$ & $|\mathcal{D}|$ & \# of Sentences\\
    \hline
    \textbf{MAG-CS} & 24,754 & 42,329 & 6 & 227,609\\
    \textbf{MAG-PSY} & 23,187 & 30,041 & 6& 111,194\\
    \textbf{WordNet-Noun} & 83,073 & 76,812 & 20 & 236,454\\
    \textbf{WordNet-Verb} & 13,936 & 13,403 & 13 & 34,654\\
    \hline
    \end{tabular}
    
    \label{data}
\end{table}
\section{Experiments}
\begin{table*}
    \caption{Overall results of Taxonomy Completion task on the four large-scale datasets. ** indicates the results are from \cite{zhang2021taxonomy}.}
  \resizebox{\linewidth}{!}{
  \begin{tabular}{c|c|c|c|c|c|c|c|c}
    \hline
    \multirow{2}{*}{\textbf{Method}} &
     \multicolumn{8}{c}{\textbf{MAG-CS}}\\
    \cline{2-9}
    & \textbf{MR} & \textbf{MRR} & \textbf{Recall@1} & \textbf{Recall@5} & \textbf{Recall@10} & \textbf{Precision@1} & \textbf{Precision@5}&\textbf{Precision@10} \\
	\cline{1-9}
	Bilinear & 3360.343 $\pm$ 6.126 & 0.026 $\pm$ 0.000 & 0.000 $\pm$ 0.000 & 0.003 $\pm$ 0.000 & 0.006 $\pm$ 0.000& 0.001$\pm$ 0.000 & 0.002$\pm$ 0.000 & 0.003$\pm$ 0.000\\
	\TaxoExpan  & 823.075 $\pm$ 114.638  & 0.193 $\pm$ 0.007 & 0.030 $\pm$ 0.002 & 0.095 $\pm$ 0.004 & 0.137 $\pm$ 0.007 & 0.132 $\pm$ 0.010 & 0.083 $\pm$ 0.003 & 0.059 $\pm$ 0.003\\
	ARBORIST**  & 1142.335 $\pm$ 19.249 & 0.133 $\pm$ 0.004 & 0.008 $\pm$ 0.001 & 0.044 $\pm$ 0.003 & 0.075 $\pm$ 0.003 & 0.037 $\pm$ 0.004 & 0.038 $\pm$ 0.003 & 0.033 $\pm$ 0.001 \\
	\TMN &  436.319 $\pm$ 13.128 & 0.243 $\pm$ 0.005 & 0.056 $\pm$ 0.001 & 0.145 $\pm$ 0.004 & 0.189 $\pm$ 0.005 & 0.245 $\pm$ 0.006  & 0.126 $\pm$ 0.003 & 0.082 $\pm$ 0.002\\
	\GenTaxo  & 13213.731 $\pm$ 662.688 & 0.239 $\pm$ 0.006 &0.082 $\pm$ 0.002 & 0.185 $\pm$ 0.008 & 0.218 $\pm$ 0.008 & 0.254 $\pm$ 0.010 & 0.131 $\pm$ 0.007 & 0.085 $\pm$ 0.003\\
	\midrule
	\TaxoEnrichS & \textbf{73.680 $\pm$ 1.346} & 0.545 $\pm$ 0.002 & 0.154 $\pm$ 0.006 & 0.396 $\pm$ 0.003 & 0.534 $\pm$ 0.002 & 0.251 $\pm$ 0.016 & 0.129 $\pm$ 0.002 & 0.087 $\pm$ 0.001\\
	\TaxoEnrich & 87.798 $\pm$ 1.512 & \textbf{0.578 $\pm$ 0.001} & \textbf{0.162 $\pm$ 0.004} & \textbf{0.434 $\pm$ 0.005} & \textbf{0.574 $\pm$ 0.003} & \textbf{0.274 $\pm$ 0.017} & \textbf{0.141 $\pm$ 0.002} & \textbf{0.093 $\pm$ 0.002}\\
	\cline{1-9}	
    \end{tabular}
  }\\
  \resizebox{\linewidth}{!}{
  \begin{tabular}{c|c|c|c|c|c|c|c|c}
    \hline
    \multirow{2}{*}{\textbf{Method}} &
 \multicolumn{8}{c}{\textbf{MAG-PSY}}\\
    \cline{2-9}
    &\textbf{MR} & \textbf{MRR} & \textbf{Recall@1} & \textbf{Recall@5} & \textbf{Recall@10} & \textbf{Precision@1} & \textbf{Precision@5}&\textbf{Precision@10} \\
	\cline{1-9}
	Bilinear  &  2118.204 $\pm$ 4.152  & 0.032 $\pm$ 0.000 & 0.000 $\pm$ 0.000 & 0.001 $\pm$ 0.000 & 0.003 $\pm$ 0.000 & 0.000 $\pm$ 0.000 & 0.000 $\pm$ 0.000 & 0.000 $\pm$ 0.000\\
	\TaxoExpan &  345.679 $\pm$ 24.306 & 0.441 $\pm$ 0.005 & 0.122 $\pm$ 0.003 & 0.287 $\pm$ 0.007 & 0.364 $\pm$ 0.009 & 0.249 $\pm$ 0.007 & 0.117 $\pm$ 0.003 & 0.074 $\pm$ 0.002\\
	ARBORIST** &  547.723 $\pm$ 20.165 & 0.344 $\pm$ 0.012 &  0.062 $\pm$ 0.009& 0.185 $\pm$ 0.011 & 0.256 $\pm$ 0.013 & 0.126 $\pm$ 0.018 & 0.076 $\pm$ 0.004 & 0.052 $\pm$ 0.003\\
	\TMN &  159.550 $\pm$ 5.290 & 0.531 $\pm$ 0.007 & 0.175 $\pm$ 0.002  & 0.369 $\pm$ 0.005 & 0.446 $\pm$ 0.009 & 0.358 $\pm$ 0.004  & 0.150 $\pm$ 0.002 & 0.091 $\pm$ 0.002\\
	\GenTaxo &  7482.516 $\pm$ 2600.713 & 0.464 $\pm$ 0.022 & 0.183 $\pm$ 0.116 & 0.402 $\pm$ 0.066 & 0.440 $\pm$ 0.039 & 0.376 $\pm$ 0.119 &0.164 $\pm$ 0.027 & 0.090 $\pm$ 0.008\\
	\midrule
	\TaxoEnrichS &  149.660 $\pm$ 3.430 & 0.561 $\pm$ 0.005 & 0.221 $\pm$ 0.010 & 0.420 $\pm$ 0.007 & 0.480 $\pm$ 0.007 & 0.365 $\pm$ 0.020 & 0.178 $\pm$ 0.003 & 0.117 $\pm$ 0.001\\
	\TaxoEnrich  & \textbf{122.247 $\pm$ 3.241} & \textbf{0.583 $\pm$ 0.010} & \textbf{0.234 $\pm$ 0.009} & \textbf{0.424 $\pm$ 0.013} & \textbf{0.510 $\pm$ 0.018} & \textbf{0.374 $\pm$ 0.021} & \textbf{0.186 $\pm$ 0.002} & \textbf{0.124 $\pm$ 0.002} \\
	\cline{1-9}	
    \end{tabular}
  }\\
  \resizebox{\linewidth}{!}{
  \begin{tabular}{c|c|c|c|c|c|c|c|c}
    \hline
    \multirow{2}{*}{\textbf{Method}} &
 \multicolumn{8}{c}{\textbf{WordNet-Noun}}\\
    \cline{2-9}
    & \textbf{MR} & \textbf{MRR} & \textbf{Recall@1} & \textbf{Recall@5} & \textbf{Recall@10} & \textbf{Precision@1} & \textbf{Precision@5}&\textbf{Precision@10} \\
	\cline{1-9}
	Bilinear  & 3290.858 $\pm$ 14.668 & 0.196 $\pm$ 0.000 & 0.013 $\pm$ 0.000 & 0.063 $\pm$ 0.000 & 0.109 $\pm$ 0.000 & 0.023 $\pm$ 0.001  & 0.022 $\pm$ 0.000 & 0.019 $\pm$ 0.000 \\
	\TaxoExpan  & 970.858 $\pm$ 50.995 & 0.390 $\pm$ 0.004 & 0.066 $\pm$ 0.002 & 0.186 $\pm$ 0.003 & 0.269 $\pm$ 0.007 & 0.114 $\pm$ 0.003 & 0.065 $\pm$ 0.001 & 0.047 $\pm$ 0.001\\
	ARBORIST**  & 2993.341 $\pm$ 114.749 & 0.217 $\pm$ 0.005 & 0.021 $\pm$ 0.001 & 0.073 $\pm$ 0.002 & 0.125 $\pm$ 0.002 & 0.036 $\pm$ 0.021 & 0.025 $\pm$ 0.001  & 0.022 $\pm$ 0.000\\
	\TMN &  827.371 $\pm$ 24.310 & 0.367 $\pm$ 0.006 & 0.054 $\pm$ 0.002 & 0.169 $\pm$ 0.002 & 0.256 $\pm$ 0.004  & 0.095 $\pm$ 0.002 & 0.058 $\pm$ 0.000 & 0.044 $\pm$ 0.001\\
	\GenTaxo  & 57871.589 $\pm$ 89.230 & 0.286 $\pm$ 0.162 &0.025 $\pm$ 0.007 & 0.169 $\pm$ 0.049 & 0.268 $\pm$ 0.118 & 0.109 $\pm$ 0.013 & 0.024 $\pm$ 0.007 & 0.029 $\pm$ 0.001\\
	\midrule
	\TaxoEnrichS  & 230.576 $\pm$ 6.472 & 0.426 $\pm$ 0.018 & \textbf{0.125 $\pm$ 0.019}  & 0.212 $\pm$ 0.012 & 0.321 $\pm$ 0.018 & 0.216 $\pm$ 0.024 & 0.108 $\pm$ 0.004 & 0.078 $\pm$ 0.003\\
	\TaxoEnrich  & \textbf{227.839 $\pm$ 12.247} & \textbf{0.442 $\pm$ 0.018} & 0.123 $\pm$ 0.012  & \textbf{0.248 $\pm$ 0.011} & \textbf{0.351 $\pm$ 0.019} & \textbf{0.226 $\pm$ 0.023} & \textbf{0.115 $\pm$ 0.002} & \textbf{0.098 $\pm$ 0.002}\\
	\cline{1-9}	
    \end{tabular}
  }\\
  \resizebox{\linewidth}{!}{
  \begin{tabular}{c|c|c|c|c|c|c|c|c}
    \hline
    \multirow{2}{*}{\textbf{Method}} &
 \multicolumn{8}{c}{\textbf{WordNet-Verb}}\\
    \cline{2-9}
    &\textbf{MR} & \textbf{MRR} & \textbf{Recall@1} & \textbf{Recall@5} & \textbf{Recall@10} & \textbf{Precision@1} & \textbf{Precision@5}&\textbf{Precision@10} \\
	\cline{1-9}
	Bilinear  &  1866.736 $\pm$ 5.020 & 0.174 $\pm$ 0.000 & 0.012 $\pm$ 0.001 & 0.054 $\pm$ 0.000 & 0.095 $\pm$ 0.000& 0.017 $\pm$ 0.001 & 0.016 $\pm$ 0.000 & 0.014 $\pm$ 0.000\\
	\TaxoExpan &  853.308 $\pm$ 18.302 & 0.325 $\pm$ 0.007 & 0.069 $\pm$ 0.001 & 0.169 $\pm$ 0.003 &0.228 $\pm$ 0.008 & 0.104 $\pm$ 0.002 & 0.051 $\pm$ 0.001 & 0.034 $\pm$ 0.001\\
	ARBORIST** &  2993.341 $\pm$ 4.950 & 0.206 $\pm$ 0.011 & 0.016 $\pm$ 0.004 & 0.073 $\pm$ 0.011 & 0.016 $\pm$ 0.011 & 0.024 $\pm$ 0.006 & 0.022 $\pm$ 0.003 & 0.018 $\pm$ 0.002\\
	\TMN &  832.541 $\pm$ 29.589 & 0.354 $\pm$ 0.010 & 0.081 $\pm$ 0.007  & 0.194 $\pm$ 0.013 & 0.259 $\pm$ 0.014 & 0.121 $\pm$ 0.011  & 0.059 $\pm$ 0.004 & 0.039 $\pm$ 0.002\\
	\GenTaxo &  2765.745  $\pm$ 262.631 & 0.428 $\pm$ 0.117 & 0.118 $\pm$ 0.069 & 0.208 $\pm$ 0.104 &  0.239 $\pm$ 0.112 & 0.235 $\pm$ 0.152 & 0.122 $\pm$ 0.038 & 0.066 $\pm$ 0.016\\
	\midrule
	\TaxoEnrichS &  \textbf{304.565} $\pm$ 3.628 & 0.442 $\pm$ 0.004 & 0.128 $\pm$ 0.003 & \textbf{0.256 $\pm$ 0.012} & \textbf{0.350 $\pm$ 0.009} & 0.242 $\pm$ 0.005 & 0.121 $\pm$ 0.004 & 0.074 $\pm$ 0.001\\
	\TaxoEnrich & 320.064 $\pm$ 14.153 & \textbf{0.452 $\pm$ 0.005} & \textbf{0.143 $\pm$ 0.002} & 0.252 $\pm$ 0.014 & 0.347 $\pm$ 0.006 & \textbf{0.276 $\pm$ 0.004} & \textbf{0.126 $\pm$ 0.001} & \textbf{0.081 $\pm$ 0.002} \\
	\cline{1-9}	
    \end{tabular}
    }\\
    
    \label{res1}

\end{table*}

\subsection{Experiment Setup}

\textbf{Dataset.} We evaluate the performance of \TaxoEnrich framework on the following four real-world large-scale datasets. The statistics of each dataset are listed in Table 1.
\begin{itemize}
    \item \textbf{Microsoft Academic Graph (MAG): } This public Field-of-Study (FoS) taxonomy contains over 660 thousand scientific concepts and more than 700 thousand taxonomic relations. We follow the data preprocessing in \cite{shen2020taxoexpan} to only select partial taxonomies under the computer science (\textbf{MAG-CS}) and psychology (\textbf{MAG-PSY}) domain \cite{sinha2015overview}. 
    \item \textbf{WordNet:} We collect the concepts and taxonomic relations from verbs and nouns sub-taxonomies based on WordNet 3.0 (\textbf{WordNet-Noun}, \textbf{WordNet-Verb}). These two sub-fields are the only parts that have fully-developed taxonomies in WordNet. 
    In practice, due to the scarcity in the dataset, i.e. there are many disconnected components in the both taxonomies, we added a pseudo root named \mquote{Noun} and \mquote{Verb} and connect this root to the head of each connected components in the taxonomies for generate a more complete taxonomic structure.
\end{itemize}
Follow the dataset splitting settings used in \cite{shen2020taxoexpan, zhang2021taxonomy}, we sample 1000 nodes for validation and test respectively in each dataset. Then we use the remaining nodes to construct the initial taxonomy. 
\subsection{Compared Methods}
To fully understand the performance of our framework, we compare our model with the following methods.
\begin{itemize}
    \item \textbf{Bilinear Model} \cite{sutskever2009modelling} incorporates the interaction of two concept embeddings, i.e., $\langle$parent, child$\rangle$ entity pair embeddings, through a simple bilinear form. This method serves as a baseline result to check the comparable performance of each framework.
    \item {\sf \textbf{TaxoExpan} } \cite{shen2020taxoexpan} is a state-of-the-art taxonomy expansion framework, which leverages the positional-enhanced graph neural network to capture the relationship between query nodes and local egonet, along with InfoNCE loss \cite{oord2018representation} to increase the robustness of the model.
    \item \textbf{ARBORIST} \cite{manzoor2020expanding} is a state-of-the-art taxonomy expansion framework which aims for taxonomies with heterogeneous edge semantics and optimizes a large margin ranking loss with a dynamic margin function.
    \item {\sf \textbf{TMN} }\cite{zhang2021taxonomy} is a state-of-the-art taxonomy completion framework and also the first framework that proposed the completion task, and computed the matching score between the query concept and $\langle$ hypernym, hyponym$\rangle$ pairs.
    \item {\sf \textbf{GenTaxo}} \cite{zeng2021enhancing} is a state-of-the-art taxonomy completion framework using both sentence-based and subgraph-based encodings of the nodes to perform the matching. Since part of the framework concentrates on concept name generation tasks, which is not the focus of this paper, we adopt the {\sf \textbf{ GenTaxo++} } assuming the newly added nodes are given. \footnote{Note that since the implementation code of {\sf GenTaxo} \cite{zeng2021enhancing} is not released, we implemented the framework based on the description in the paper.}
\end{itemize}

We also include two variants of \TaxoEnrich in experiments for ablation study:

\begin{itemize}
    \item {\sf{\textbf{TaxoEnrich-S}}}: In this version, we exclude the sibling information from the matching model, since in sparse taxonomies, such as WordNet, the siblings cannot represent the precise candidate positions, and might still introduce noisy information when computing the relateness between query node and candidate position.
    \item {\sf{\textbf{TaxoEnrich}}}: In this version, we adopt the full framework of \TaxoEnrich as described above. We will examine the difference between two variants through further experiments.
\end{itemize}

\subsection{Evaluation Metrics}
Since the result from the model's output is a ranking list of candidate positions for each query node, following the guidelines in \cite{shen2020taxoexpan, zhang2021taxonomy}, we utilize the following rank-based metrics to evaluate the performance our framework and the comparison methods.
\begin{itemize}
    \item \textbf{Mean Rank (MR)}. This metric measures the average rank position of a query concept's true position among all candidate positions. For queries with multiple correct positions, we first calculate the rank position of each individual triplet and then take the average of all rank positions. Smaller value in this metric indicates the better performance of the model.
    \item \textbf{Mean Reciprocal Rank (MRR)}. We follow \cite{ying2018graph} to compute the reciprocal rank of a query concept's true positions using a scaled MRR. In the evaluation, we scale the MRR by 10 to enlarge the difference between different models clearly.
    \item \textbf{Recall@}$k$ measures the number of query concepts’ true positions ranked in the top $k$, divided by the total number of true positions of all query concepts.
    \item \textbf{Precision@}$k$ measures the number of query concepts’ true positions ranked in the top $k$, divided by the total number of queries times $k$.
\end{itemize}
For all the evaluation metrics listed above except for MR, the larger value indicates better performance of the model. During the evaluation, since MR and MRR are the only metrics that concentrates on the performance of all predictions in the taxonomy in general, we consider them as the most important metric for evaluation.

\section{Experimental Results}
In this section we will first discuss the experiment results on both taxonomy completion and expansion tasks which demonstrated the superiority of our \TaxoEnrich method. Then to further understand the contributions from each of our model design, we conduct ablation studies. Finally we performed case studies to further illustrate the effectiveness of {\sf TaxoEnrich}.
\subsection{Performance on Taxonomy Completion}
The overall performance of compared methods and the proposed framework is indicated in Table \ref{res1}. First, we can see that the performance of the framework tends to become better when the complexity of local structure increases, from the one-to-one matching in \TaxoExpan to triplet in {\sf \TMN}, and the neighboring paths and subgraph encoding in {\sf GenTaxo}. Second, we can generally observe the power of pre-trained language models in the representations of concept nodes in the taxonomy. The frameworks including {\sf GenTaxo} and \TaxoEnrich utilizing language models have generally better performance in the precision@$k$ and recall@$k$ metrics. 

In terms of MR , we can see that \TaxoEnrich obtained most performance improvement in MAG-CS dataset since the computer science taxonomy has the most complete taxonomic structure compared with other datasets, allowing for more accurate taxonomy-contextualized embeddings generated by Section \ref{31}. And in WordNet datasets the performance in MR metric is improved by a relatively large margin while all frameworks do not perform as well as in MAG datasets. In terms of precision@$k$ and recall@$k$, our method also shows noticeable improvement over baseline models. In the previous methods, the static embedding method failed to capture the similar semantic meaning between different concept nodes. And we can see {\sf GenTaxo} renders competing performance on these two metrics, but tends to be unstable and perform not well in ranking metrics. The primary reason for this observation is that while the language-based embeddings can provide pretty accurate positional information, its light-weight MLP matching module prevents it from capturing useful relatedness between query node and candidate position.

For two WordNet datasets, we can see that other frameworks are inclined to have similarly poor performance due to the scarcity of taxonomies. The non-connectivity causes the matching module difficult to extract the relations during the training. Therefore, the manually added pseudo root for sentence generation would maintain the taxonomic structure information in the representations of concept node and candidate positions, allowing the framework to capture both the structure and semantic information of each node.

In the comparison between {\sf TaxoEnrich-S} and {\sf TaxoEnrich}, we can observe that, in two MAG datasets, the incorporation of sibling information in {\sf TaxoEnrich} would have better performance. However, it will also cause a drop in MR metric except for MAG-PSY and WordNet-Noun datasets since the randomly extracted siblings will still introduce noisy information in the matching module. In the WordNet datasets, the performance of two methods are very similar. This is because with the scarcity in the taxonomies, i.e., the lack of siblings, will mislead the model to incorporate inaccurate sibling information, causing a clear difference for MR metric. On the other hand, the precision in {\sf TaxoEnrich} is still better than {\sf TaxoEnrich-S}, which illustrates the effectiveness of siblings in representing the positional information.

\subsection{Performance on Taxonomy Expansion}
Taxonomy expansion is a special case of the taxonomy completion task where new concepts are all leaf nodes. In this case, we would like to further explore the performance of \TaxoEnrich on taxonomy expansion task on MAG-CS and WordNet-Verb dataset, compared with {\sf TaxoExpan}, {\sf TMN}, and {\sf GenTaxo} framework. As indicated in Table \ref{res2}, we can also observe that \TaxoEnrich outperforms other methods by a large margin in all metrics.
\begin{table}[ht]
    \caption{Results of Taxonomy Expansion task on the MAG-CS and WordNet-Verb datasets.}
    \resizebox{\linewidth}{!}{
        \begin{tabular}{|c|c|c|c|c|}
        \hline
            \multirow{2}{*}{\textbf{Method}} &
            \multicolumn{4}{|c|}{\textbf{MAG-CS}}\\
        \cline{2-5}
            &\textbf{MR} & \textbf{MRR} &\textbf{Recall@1} &  \textbf{Precision@1} \\
        \cline{1-5}
            \TaxoExpan  & 197.776 $\pm$ 16.038 & 0.562 $\pm$ 0.023 & 0.100 $\pm$ 0.011 & 0.163 $\pm$ 0.018\\
            \TMN & 118.963 $\pm$ 6.307 & 0.689 $\pm$ 0.005 & 0.174 $\pm$ 0.002 & 0.283 $\pm$ 0.004\\
            \GenTaxo & 140.262 $\pm$ 40.398 & 0.634 $\pm$ 0.044 & 0.149 $\pm$ 0.020 & 0.294 $\pm$ 0.096 \\
            \midrule
            \TaxoEnrichS  & \textbf{67.947 $\pm$ 1.121} & \textbf{0.721 $\pm$ 0.008} &  \textbf{0.182 $\pm$ 0.005} & \textbf{0.304 $\pm$ 0.008}\\
        \cline{1-5}
        \end{tabular}
    }\\
    \resizebox{\linewidth}{!}{
        \begin{tabular}{|c|c|c|c|c|}
            \hline
                \multirow{2}{*}{\textbf{Method}} &
                \multicolumn{4}{|c|}{\textbf{WordNet-Verb}}\\
            \cline{2-5}
                &\textbf{MR} & \textbf{MRR} & \textbf{Recall@1} &  \textbf{Precision@1} \\
            \cline{1-5}
                \TaxoExpan  & 665.409 $\pm$ 137.250 & 0.406 $\pm$ 0.056 &  0.085 $\pm$ 0.018& 0.095 $\pm$ 0.004\\
                \TMN & 615.021 $\pm$ 166.375 &  0.423 $\pm$ 0.056 & 0.110 $\pm$ 0.021 & 0.124 $\pm$ 0.009\\
                \GenTaxo  & 6046.363 $\pm$ 439.305 & 0.155 $\pm$ 0.010 & 0.094 $\pm$ 0.019 & 0.141 $\pm$ 0.079\\
                \midrule
                \TaxoEnrichS & \textbf{217.842  $\pm$ 5.230} & \textbf{0.481 $\pm$ 0.071} & \textbf{0.162 $\pm$ 0.082} & \textbf{0.294 $\pm$ 0.031}\\
        \cline{1-5}
        \end{tabular}
    }\\
    \label{res2}
\end{table}

\begin{table}[ht]
    \caption{Ablation studies on the incorporation of siblings in embeddings on MAG-CS dataset. The framework that randomly incorporates sibling information in embedding generation module is denoted as {\sf TaxoEnrich-Sib} for simplicity.}
    \label{res3}
    \centering
    \resizebox{\linewidth}{!}{
    
    \begin{tabular}{|c|c|c|c|c|}
    \hline
    \multirow{2}{*}{\textbf{Method}} &
    \multicolumn{4}{|c|}{\textbf{MAG-CS}}\\
    \cline{2-5}
    &\textbf{MR} & \textbf{MRR} &\textbf{Recall@1} &  \textbf{Precision@1} \\
    \cline{1-5}
    {\sf TaxoEnrich-Sib}  & 122.144 $\pm$ 3.219 & 0.513 $\pm$ 0.006 &  0.138 $\pm$ 0.000 & 0.224 $\pm$ 0.001\\
    \TaxoEnrichS  & \textbf{73.680 $\pm$ 1.346} & \textbf{0.545 $\pm$ 0.002} & \textbf{0.154 $\pm$ 0.006} & \textbf{0.251 $\pm$ 0.016}\\
    \cline{1-5}
    \end{tabular}
    }

\end{table}

\begin{table}
    \centering
    \caption{Ablation studies on MAG-CS dataset with different feature encoders. Some results are from the main table.}
    \resizebox{\linewidth}{!}{
    
  \begin{tabular}{|c|c|c|c|c|}
    \hline
    \multirow{2}{*}{\textbf{Method}} & \multirow{2}{*}{\textbf{Distribution Model}} &
    \multicolumn{3}{|c|}{\textbf{MAG-CS}}\\
    \cline{3-5}
    & & \textbf{MR}  &\textbf{Recall@1} &  \textbf{Precision@1} \\
    \cline{1-5}
	\TaxoExpan & Raw Embedding  & 3360.343 $\pm$ 6.126 &  0.000 $\pm$ 0.000 & 0.001 $\pm$ 0.001\\
	\TaxoExpan & Raw + PGAT  & 823.075 $\pm$ 114.638  &  0.030 $\pm$ 0.002 & 0.132 $\pm$ 0.010\\
	\TMN &  Raw Embedding & 636.254 $\pm$ 36.465 &  0.036 $\pm$ 0.005  & 0.156 $\pm$ 0.008\\
	\TMN & Raw + LSTM + PGAT & 436.319 $\pm$ 13.128 &  0.056 $\pm$ 0.001 & 0.245 $\pm$ 0.006\\
	\midrule
	\TaxoEnrichS & Raw Embedding &  103.016 $\pm$ 6.589 & 0.145 $\pm$ 0.004 & 0.236 $\pm$ 0.010\\
	\TaxoEnrichS & Raw + LSTM  & \textbf{73.680 $\pm$ 1.346} & \textbf{0.154 $\pm$ 0.006} & \textbf{0.251 $\pm$ 0.024}\\
	\TaxoEnrichS & Raw + LSTM + PGAT  &  100.188 $\pm$ 2.214 &  0.150 $\pm$ 0.004 & 0.244 $\pm$ 0.001\\
	\cline{1-5}
    \end{tabular}
    }
    
    \label{res4}
\end{table}

\begin{table*}[ht]
      \centering
      \caption{Case Studies of predicted positions on MAG-CS with both leaf and internal query concepts}
    \resizebox{0.95\linewidth}{!}{

  \begin{tabular}{|c|c|c|c|}
    \hline
    \multirow{3}{*}{\textbf{Method}}  &
    \multicolumn{3}{|c|}{\textbf{MAG-CS}}\\
    \cline{2-4}
    & \multirow{2}{*}{\textbf{Query}}  & \multirow{2}{*}{\textbf{Top 5 Predicted Positions $\binom{\text{hypernym}}{\text{hyponym}}$, True Positions in \textcolor{red}{Red}}} &  \multirow{2}{*}{\textbf{Ranking of True Positions}} \\
    & & &\\
    \cline{1-4}
    \multirow{2}{*}{\TaxoEnrich} & \multirow{4}{*}{heap} & \multirow{2}{*}{
    $\binom{\text{\textcolor{red}{programming language}}}{\text{\textcolor{red}{heaps law} }}, \binom{\text{programming language}}{\text{pseudo leaf}}, \binom{\text{algorithm}}{\text{pseudo leaf}}, \binom{\text{\textcolor{red}{programming language}}}{\text{\textcolor{red}{heap overflow}}},
    \binom{\text{\textcolor{red}{algorithm}}}{\text{\textcolor{red}{heap overflow}}}$
    } & \multirow{2}{*}{1,4,5}\\
	& & &\\
	\cline{1-1}\cline{3-4}
	\multirow{2}{*}{\TMN} &  & \multirow{2}{*}{
    $\binom{\text{reference counting}}{\text{pseudo leaf}}, \binom{\text{garbage collection}}{\text{pseudo leaf}}, \binom{\text{programming language}}{\text{pseudo leaf}}, \binom{\text{algorithm}}{\text{pseudo leaf}},
    \binom{\text{binary heap}}{\text{collection}}$
    } & \multirow{2}{*}{530, 634, 4884}\\
	& & &\\
	\cline{1-4}
    \multirow{2}{*}{\TaxoEnrich} & \multirow{4}{*}{perl} & \multirow{2}{*}{
    $\binom{\text{programming language}}{\text{pseudo leaf}}, \binom{\text{operating system}}{\text{pseudo leaf}}, \binom{\text{\textcolor{red}{programming language}}}{\text{\textcolor{red}{perl critic}}}, \binom{\text{\textcolor{red}{programming language}}}{\text{\textcolor{red}{perl 6}}},
    \binom{\text{programming language}}{\text{heap lustre}}$
    } & \multirow{2}{*}{2,3}\\
	& & &\\
	\cline{1-1}\cline{3-4}
	
	\multirow{2}{*}{\TMN} &  & \multirow{2}{*}{
    $\binom{\text{programming language}}{\text{pseudo leaf}}, \binom{\text{software}}{\text{pseudo leaf}}, \binom{\text{operating system}}{\text{pseudo leaf}}, \binom{\text{database}}{\text{pseudo leaf}},
    \binom{\text{scripting language}}{\text{perl critic}}$
    } & \multirow{2}{*}{12, 345}\\
	& & &\\
	\cline{1-4}
	\multirow{2}{*}{\TaxoEnrich} & \multirow{4}{*}{all pairs testing} & \multirow{2}{*}{
    $\binom{\text{\textcolor{red}{test suite}}}{\text{\textcolor{red}{pseudo leaf}}}, \binom{\text{\textcolor{red}{model based testing}}}{\text{\textcolor{red}{pseudo leaf}}}, \binom{\text{metamorphic testing}}{\text{pseudo leaf}}, \binom{\text{test driven development}}{\text{pseudo leaf}},
    \binom{\text{redundant code}}{\text{perl critic}}$
    } & \multirow{2}{*}{1, 2}\\
	& & &\\
	\cline{1-1}\cline{3-4}
	\multirow{2}{*}{\TMN} &  & \multirow{2}{*}{
    $\binom{\text{metamorphic testing}}{\text{pseudo leaf}}, \binom{\text{random testing}}{\text{pseudo leaf}}, \binom{\text{artificial intelligence}}{\text{pseudo leaf}}, \binom{\text{algorithm}}{\text{pseudo leaf}},
    \binom{\text{machine learning}}{\text{pseudo leaf}}$
    } & \multirow{2}{*}{6, 133}\\
	& & &\\
	\cline{1-4}
	\multirow{2}{*}{\TaxoEnrich} & \multirow{4}{*}{sensor hub} & \multirow{2}{*}{
    $\binom{\text{\textcolor{red}{embedded system}}}{\text{\textcolor{red}{pseudo leaf}}}, \binom{\text{operating system}}{\text{pseudo leaf}}, \binom{\text{\textcolor{red}{computer network}}}{\text{\textcolor{red}{pseudo leaf}}}, \binom{\text{\textcolor{red}{computer hardware}}}{\text{\textcolor{red}{pseudo leaf}}},
    \binom{\text{embedded system}}{\text{virtual metrology}}$
    } & \multirow{2}{*}{1, 3, 4}\\
	& & &\\
	\cline{1-1}\cline{3-4}
	\multirow{2}{*}{\TMN} & & \multirow{2}{*}{
    $\binom{\text{\textcolor{red}{embedded system}}}{\text{\textcolor{red}{pseudo leaf}}}, \binom{\text{operating system}}{\text{pseudo leaf}}, \binom{\text{\textcolor{red}{computer hardware}}}{\text{\textcolor{red}{pseudo leaf}}}, \binom{\text{software}}{\text{pseudo leaf}},
    \binom{\text{wearable computer}}{\text{pseudo leaf}}$
    } & \multirow{2}{*}{1, 78, 3}\\
	& & &\\
	\cline{1-4}
    \end{tabular}
    }
    
    \label{case}
\end{table*}
\subsection{Ablation Studies}
In this section, we conduct the ablation studies on the major components of \TaxoEnrich framework: 1) Incorporation of sibling information separated from embedding generation; 2) The implementations of different feature encoder models to capture different structural information. Note that, in the ablation study experiment, we used \TaxoEnrichS model for more direct and simpler comparison between the embeddings and modules. Additional ablation studies about hyperparameters are presented in appendix.

\subsubsection{The Effectiveness of Query-Aware Sibling Encoder.}
In this section, we further discuss the effectiveness of approaches of incorporating sibling information in our framework. We argue that simple including sibling information in embeddings would actually introduce noisy information to the framework. In many cases, some high-level concepts, such as \mquote{Artificial Intelligence} or \mquote{Machine Learning} in MAG-CS taxonomy, have thousands of children. Therefore, it is unrealistic to consider all siblings, or unreasonable to randomly consider some of them. Thus, we conduct experiments to verify this assumption by randomly selecting at most 5 siblings in the process of embedding generation. However, as shown in Table \ref{res3}, such operation would not only prevent the framework from recognizing correct positions, but increase the embedding generation time to the three times of the original in the implementation.

\subsubsection{Different Feature Encoders in \TaxoEnrich}
We will continue to examine the superior performance of \TaxoEnrich with different feature encoders and the effectiveness of such methods on other frameworks. The techniques of encoding features before matching have been experimented by previous methods \cite{shen2020taxoexpan, yu2020steam}, showing that the neighboring terms of the candidate position would better utilize the structural information. We implement different feature encoding: the raw embeddings, LSTM encoding, PGAT encoding, and the combinations of these three, i.e., concatenating the output of encoders as input for matching module, for comparison in this section. Through experiments in Table \ref{res4}, we can see that, the encoded features will improve the performance of all frameworks by a large margin, and \TaxoEnrich can still outperform other methods regardless of encoded embeddings of concept nodes.

\subsection{Case Studies}
We demonstrate the effectiveness of \TaxoEnrich framework by predicting true positions of several query concepts in MAG-CS datasets in Table \ref{case}. For high-level concepts like \mquote{heap}, \TaxoEnrich ranks all the true positions at top 5, while \TMN can only identify part of the true positions' information, like (\mquote{algorithm, leaf}) for \mquote{heap}. And for those leaf nodes, such as \mquote{all pairs testing} and \mquote{sensor hub}, we can observe that \TMN will have much better performance. However, it will include some coarse high-level concepts such as \mquote{machine learning} and \mquote{artificial intelligence}. In general, we can see \TaxoEnrich works better than baselines for recovering true positions, and the top predictions by \TaxoEnrich generally follow reasonable consistency.
\section{Related Work}\label{sec:related_work}

Automatic taxonomy construction is a long-standing task in the literature. Existing taxonomy construction methods leverage lexical features from the resource corpus such as lexical-patterns \cite{Nakashole2012PATTYAT,Jiang2017MetaPADMP, hearst1992automatic,Agichtein2000SnowballER} or distributional representations \cite{mao2018end, Zhang2018TaxoGenCT, Jin2018JunctionTV,Luu2016LearningTE,Roller2014InclusiveYS,Weeds2004CharacterisingMO} to construct a taxonomy from scratch. However, in many real-world applications, some existing taxonomies may have already been laboriously curated and are deployed in online systems, which calls for solutions to the taxonomy expansion problem. To this end, multitudinous methods have been proposed recently to solve the taxonomy expansion problem \cite{manzoor2020expanding, shen2020taxoexpan, yu2020steam, mao2020}. For example, 
ARBORIST \cite{manzoor2020expanding} studies expanding taxonomies by jointly learning latent representations for edge semantics and taxonomy concepts;
\TaxoExpan \cite{shen2020taxoexpan} proposes position-enhanced graph neural networks to encode the relative position of terms and a robust InfoNCE loss \cite{oord2018representation};
{\sf STEAM} \cite{yu2020steam} re-formulates the taxonomy expansion task as a mini-path-based prediction task and proposes to solve it through a multi-view co-training objective. Some other methods were proposed for taxonomy completion, such as \TMN \cite{zhang2021taxonomy} focuses on taxonomy completion task with channel-gating mechanism and triplet matching network; and {\sf GenTaxo} \cite{zeng2021enhancing} collects information from complex local-structure information and learns to generate concept's full name from corpus.

\section{Conclusion}
In this paper, we proposed \TaxoEnrich to enhance taxonomy completion task with self-supervision. It captures the hierarchical and semantic information of concept nodes based on the taxonomic relations in the existing taxonomy. Additionally, the selective query-aware attention module and elaborately designed matching module further improves the performance of learning relatedness between query node and candidate position. Extensive experimental results elucidated the effectiveness of \TaxoEnrich by showing that it largely outperforms the previous methods achieving state-of-the-art performance on both taxonomy completion and expansion tasks. 

\bibliographystyle{ACM-Reference-Format}
\bibliography{acm}

\label{appendix}
\appendix

\section{Implementation Details}
\subsection{Baseline Models}
{\sf TaxoExpan, ARBORIST} \cite{shen2020taxoexpan, manzoor2020expanding} were designed for taxonomy expansion task. We follow the implementations in \cite{zhang2021taxonomy} to calculate the ranking of candidate positions from the single score output of their matching model, so that we can have similar output for evaluations. For {\sf TaxoExpan} \cite{shen2020taxoexpan}, we implemented the full framework with PGAT propagation method and InfoNCE Loss \cite{oord2018representation}. In comparison experiments in \cite{zhang2021taxonomy}, all methods only leveraged the initial embeddings without any distribution models for matching model comparison. For \TMN, we implemented with Raw embedding + LSTM and PGAT encoders for the full comparison, based on the original triplet matching network. And for {\sf GenTaxo}, we used the same distribution model as in {\sf TaxoEnrich}.

Note that in the previous methods, such as {\sf TaxoExpan, ARBORIST} and \TMN, the generation of the initial embeddings was from the static word embedding method. In MAG datasets, the embeddings of each concept node is computed using Word2Vec method to generate a 250-dimensional vectors. And in WordNet datasets, the embeddings were generated using FastText as a 300-dimensional vector.

\subsection{Hyperparameter Settings}
In the implementation of {\sf TaxoEnrich}, we use Adam optimizer \cite{kingma2017adam} with learning rate 0.001. We applied a scheduler which multiplies the learning rate by a factor of 0.5 after 10 epochs of non-improving metrics. The hidden dimension for LSTM encoders is set as 500 and the number of bilinear models $k$ in the matching module is 10. The number of siblings selected in the attention module $t$ is set as 5 for training and 20 for testing. The model is then trained with 200 epochs with early stop if the MR metric has not improved for more than 10 epochs on validation dataset. For other hyperparameters, we set $l_1, l_2, l_3, l_4 = 1.0, 1.0, 1.0, 0.2$ in all datasets to avoid heavy parameter tuning, batch size as 16 for both \TaxoEnrich and \TaxoEnrichS.

\section{Additional Ablation Studies}
\subsection{Hyperparameter Tuning}
We conduct additional ablation studies on hyperparameter searching. We examine the influence of batch size of \TaxoEnrichS framework. It turns out that the batch size with 16 tends to be better than others.\\
\includegraphics[width=0.5\textwidth]{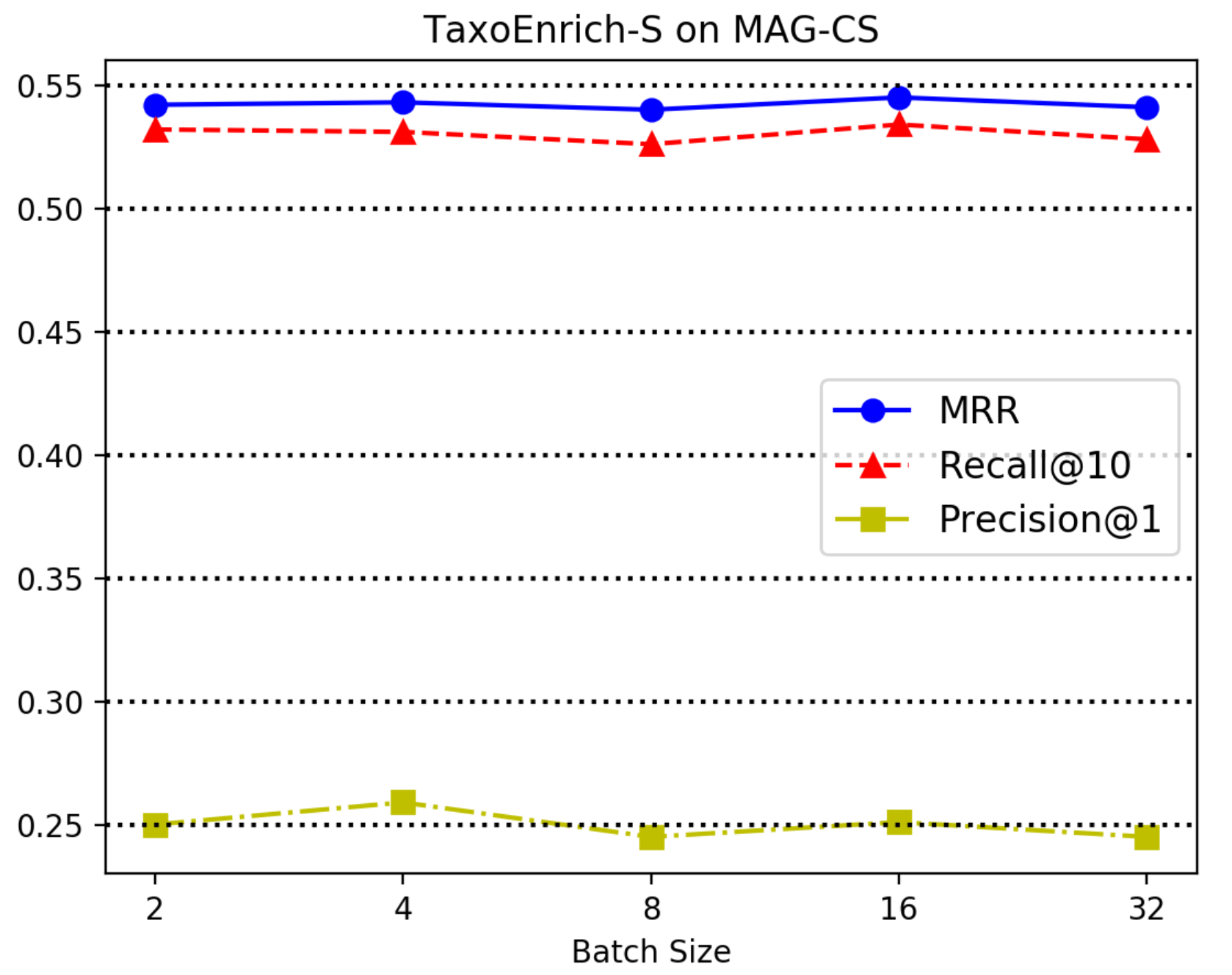}\\
Experiments on the learning rate of the newly incorporated sibling loss is also explored. We can observe that $l_4 = 0.2$ and $0.5$ will result in slightly better performance for MAG-CS datasets, as the information in siblings will still introduce noises if we treat is equally with parent and children relatedness. \\
\includegraphics[width=0.5\textwidth]{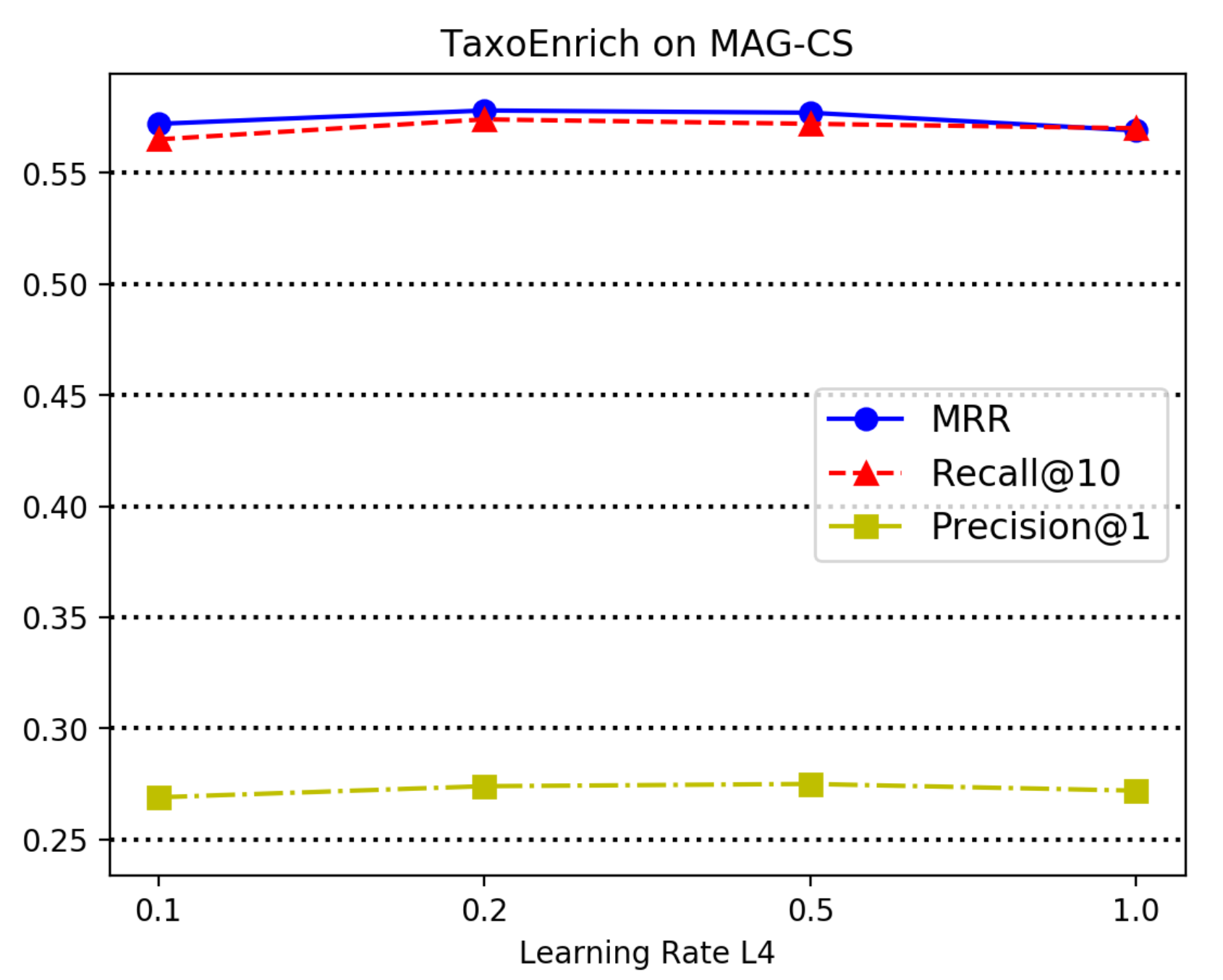}

\subsection{Sentence Encoder Studies}
The comparison between different pretrained language models for sentence encoders is also studied under settings described above. And the results are shown in Table \ref{encoder}. We can see that SciBERT achieves the best performance among all language models, and Transformer has very similar results. And BERT has relatively poor performance. The reason may be that BERT is pretrained on general domain, making it less accurate in representing scientific domain-specific concepts in MAG-CS datasets.
\begin{table}[ht]
    \caption{Ablation studies of \TaxoEnrichS on MAG-CS for the comparison between different pseudo sentence encoders.}
    \label{encoder}
    \centering
    \resizebox{\linewidth}{!}{
    
    \begin{tabular}{|c|c|c|c|}
    \hline
    \multirow{2}{*}{\textbf{Method}} &
    \multicolumn{3}{|c|}{\textbf{MAG-CS}}\\
    \cline{2-4}
    &\textbf{MR} & \textbf{Recall@1} &  \textbf{Precision@1} \\
    \cline{1-4}
    Transformer  & 74.132 &  0.149  & 0.248 \\
    SciBERT & \textbf{73.680} &  \textbf{0.154 } & \textbf{0.251 }\\
    BERT & 253.221   &  0.082  & 0.173\\
    \cline{1-4}
    \end{tabular}
    }

\end{table}
\end{spacing}
\end{document}